\documentclass[a4paper]{article}

\usepackage[english]{babel}
\usepackage[utf8x]{inputenc}
\usepackage[T1]{fontenc}

\usepackage[a4paper,top=3cm,bottom=2cm,left=3cm,right=3cm,marginparwidth=1.75cm]{geometry}

\usepackage{amsmath}
\usepackage{graphicx}
\usepackage[colorinlistoftodos]{todonotes}
\usepackage[colorlinks=true, allcolors=blue]{hyperref}

\title{Forecasting User Interests Through Topic Tag Predictions in Online Health Communities}
\date{}
\author{Amogh Subbakrishna Adishesha, Lily Jakielaszek, Fariha Azhar,\\ Peixuan
Zhang, Vasant Honavar, Fenglong Ma,\\ Chandra Belani, Prasenjit Mitra, and Sharon Xiaolei Huang}

\begin{document}
\maketitle

\begin{abstract}
The increasing reliance  on online communities for healthcare information by patients and caregivers has led to the increase in the spread of misinformation, or subjective, anecdotal and inaccurate or non-specific recommendations, which, if acted on, could cause serious harm to the patients.  Hence, there is an urgent need to connect users with accurate and tailored health information in a timely manner to prevent such harm. This paper proposes an innovative approach to suggesting reliable information to participants in online communities as they move through different stages in their disease or treatment. We hypothesize that patients with similar histories of disease progression or course of treatment would have similar information needs at comparable stages. Specifically, we pose the problem of predicting topic tags or keywords that describe the future information needs of users based on their profiles, traces of their online interactions within the community (past posts, replies) and the profiles and traces of online interactions of other users with similar profiles and similar traces of past interaction with the target users.  The result is a variant of the collaborative information filtering or recommendation system tailored to the needs of users of online health communities.  We report results of our experiments on an expert curated data set which demonstrate the superiority of the proposed approach over the state of the art baselines with respect to accurate and timely prediction of topic tags (and hence information sources of interest).
\end{abstract}


\section{Introduction}
\label{sec:introduction}
Personal healthcare management increasingly requires accommodating a higher degree of involvement of patients and informal caregivers. Consequently, patients and their caregivers expect healthcare providers to provide more detailed  information about the disease, prognosis, and treatment. However, there is a disconnect between healthcare providers and patients (and their caregivers) in terms of language used, and the information that is shared with the patients. For example, clinicians often communicate using medical terminology to describe the patient's medical condition, treatment plans, medications, medical procedures, etc. Patients have a hard time making sense of the information provided to them by their healthcare providers. The situation is not much better with information available through electronic health records (EHR)  \cite{day2019feasibility,genes2018smartphone,joukes2019impact}.

This situation is exacerbated by the fact that only 12\% of U.S. adults have proficient health literacy \cite{kutner2006health,zheng2018relationship}. Health literacy is especially low among older adults, minorities, groups with low income and socio-economic status, leading to increased healthcare usage and cost and worse health outcomes. 
Given the difficulty of comprehending health information available from healthcare providers or personal EHR, patients and caregivers increasingly turn to the internet and online communities. In such communities, patients and their caregivers interact with and seek information, e.g., regarding treatment options,  side-effects of medications, etc., from their peers, in simple, lay terms.

While online communities can be useful sources of information and emotional support for patients, they also increase the risk of misinformation \cite{morahan2000information}, or subjective \cite{zhou2019improving}, anecdotal and inaccurate or non-specific recommendations, which, if acted on by patients, could cause them serious harm.  Hence, there is an urgent need to connect users with accurate and tailored health information in a timely manner to prevent such harm.


Typically, in an online community, a user seeking information posts a natural language query; Peers, or in some cases, subject matter experts (e.g., clinicians or nurses)  respond to the query or point to useful sources of information \cite{white2014health}. Somewhat more savvy users may be able to hone in on appropriate keywords or search tokens (tags) to use in order to seek relevant information on the web (Google, Bing, etc.), medical information boards (WebMD, Mayo, etc.) or specialized platforms (HealthUnlocked, PatientsLikeMe, etc.). The keywords used in such queries are suggestive of the respective user's information needs. However, due to low health literacy, users may not necessarily be able to identify the right keywords to use; and it can be tedious and often frustrating for patients to sift through the results retrieved in response to their queries. Against this background, we explore a solution that aims to lower the barriers for patients to obtain accurate and understandable sources, e.g., articles written for laypersons that meet their health information needs.  

\begin{figure}
    \centering
    \includegraphics[width=1\columnwidth]{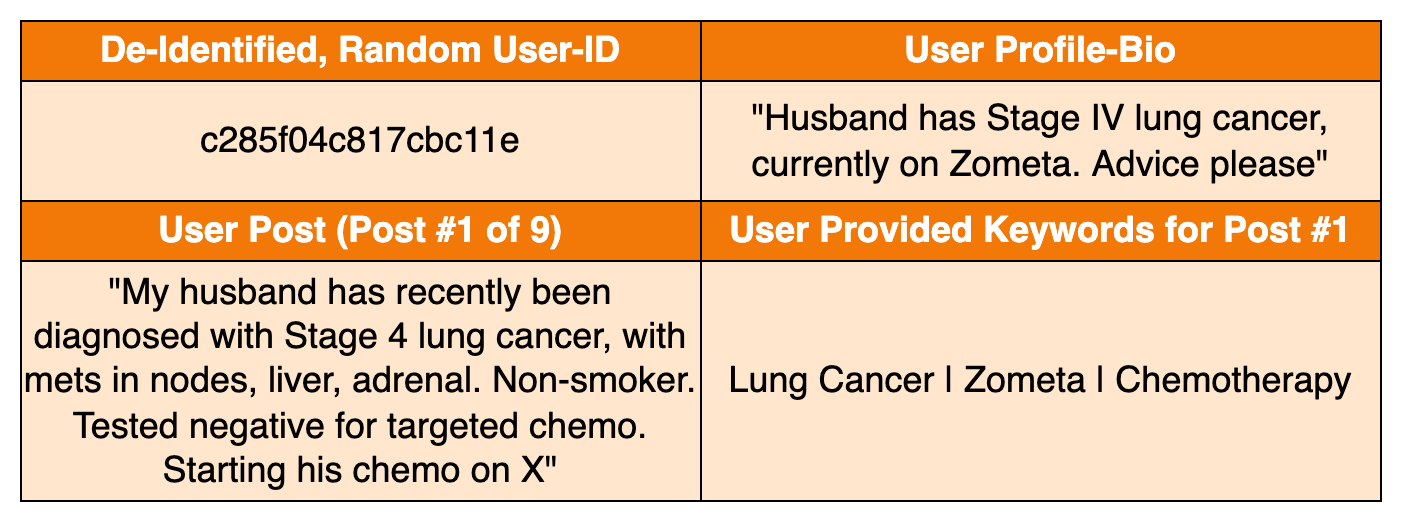}
    \caption{ Example input snippet from HealthUnlocked; The data is de-identified by HealthUnlocked where dates and names within a post are removed. The date of the post, however, is made available.}
    \label{fig:inpsnippet}
\end{figure}
In this study, we use data from  an online community HealthUnlocked\footnote{https://healthunlocked.com/} , i.e., a social network designed to offer health-related information and emotional support for patients and their caregivers. An example of the data extracted from HealthUnlocked is shown in \textbf{Figure \ref{fig:inpsnippet}}. Though \textbf{Figure \ref{fig:inpsnippet}} shows only one post, a typical user interacts a lot more with multiple posts and their corresponding keywords (tags). 
We observe that in such a setting,  information needs of a user may involve medications, side effects, treatment options and other time critical concerns. These needs depend on not only the patient demographics and health condition, but also, where the patient is with regard to progression of the disease (e.g., stage of cancer in the case of cancer patients), or  treatment being given (e.g. chemotherapy), or response to treatment (e.g., side effects, cancer remission). 

The vast amounts of healthcare information available on the internet has led to the requirement of recommendation systems that help users navigate their care by enabling them with accurate information \cite{aceto2018role} \cite{kruse2018health}.
Additionally, \cite{deng2017understanding} discusses the role of such systems in reducing risks like misinformation and time delays in access to accurate medical information. \cite{yue2021overview} reviews the recommendation models including the usage of collaborative filtering like \cite{klavsnja2018enhancing}, \cite{liao2019news} and \cite{singh2017tagme}  and finds its relevance in healthcare topic recommendation. Unlike other user-centric recommendation systems, healthcare based models are a lot harder to be subjected to collaborative filtering as there is high degree of individuality in the information sought by the users \cite{deng2019collaborative}. \cite{sahoo2019deepreco} however, proposes a CNN based feature aggregation approach to perform user-content based recommendation and presents improvements over personalization metrics. Similarly, \cite{jiang2017user} creates a similarity matching based on heterogeneous networks to perform collaborative filtering for healthcare topic recommendation.

Extracting features from text documents is at the centre of our proposed topic recommendation system. Text documents are in essence a sequence of vectors placed together with context and grammar. Early approaches to model text included the usage of recurrent neural networks (RNNs) like in \cite{guo2018cran} and \cite{nallapati2016sequence}. It became evident that the order (positions) of these vectors in relation to their value was vital in text based tasks. Bidirectional long-short term memory models (LSTMs) including \cite{kaageback2016word},\cite{melamud2016context2vec} and specifically healthcare applications like \cite{9112671} emphasized the importance of positional embeddings of the word vectors and improved performance on tasks like translation and sequence-to-sequence text generation. Transformers, a recent advancement by \cite{vaswani2017attention} along with their attention modules have revolutionised feature extraction like in \cite{wang2020detecting}. Bidirectional Encoder Representations from Transformers (BERTs) an advancement of transformers have additionally aided in text based tasks like classification \cite{garg2020bae}, \cite{luo2021deep}, summarization \cite{liu2019text}, \cite{miller2019leveraging} and text generation \cite{chan2019recurrent}, \cite{li2021pretrained}. They are ideal for sentence/token generation conditioned upon prior context \cite{9364676}. \textbf{Table \ref{table:intro}} compares our work in view of current literature. The position of current healthcare recommendation literature along with advances in natural language processing technology pose the ideal setting for a novel medical information recommendation system that predicts the future interests of users through topic tags.
\begin{figure}
    \centering
    \includegraphics[scale=0.166]{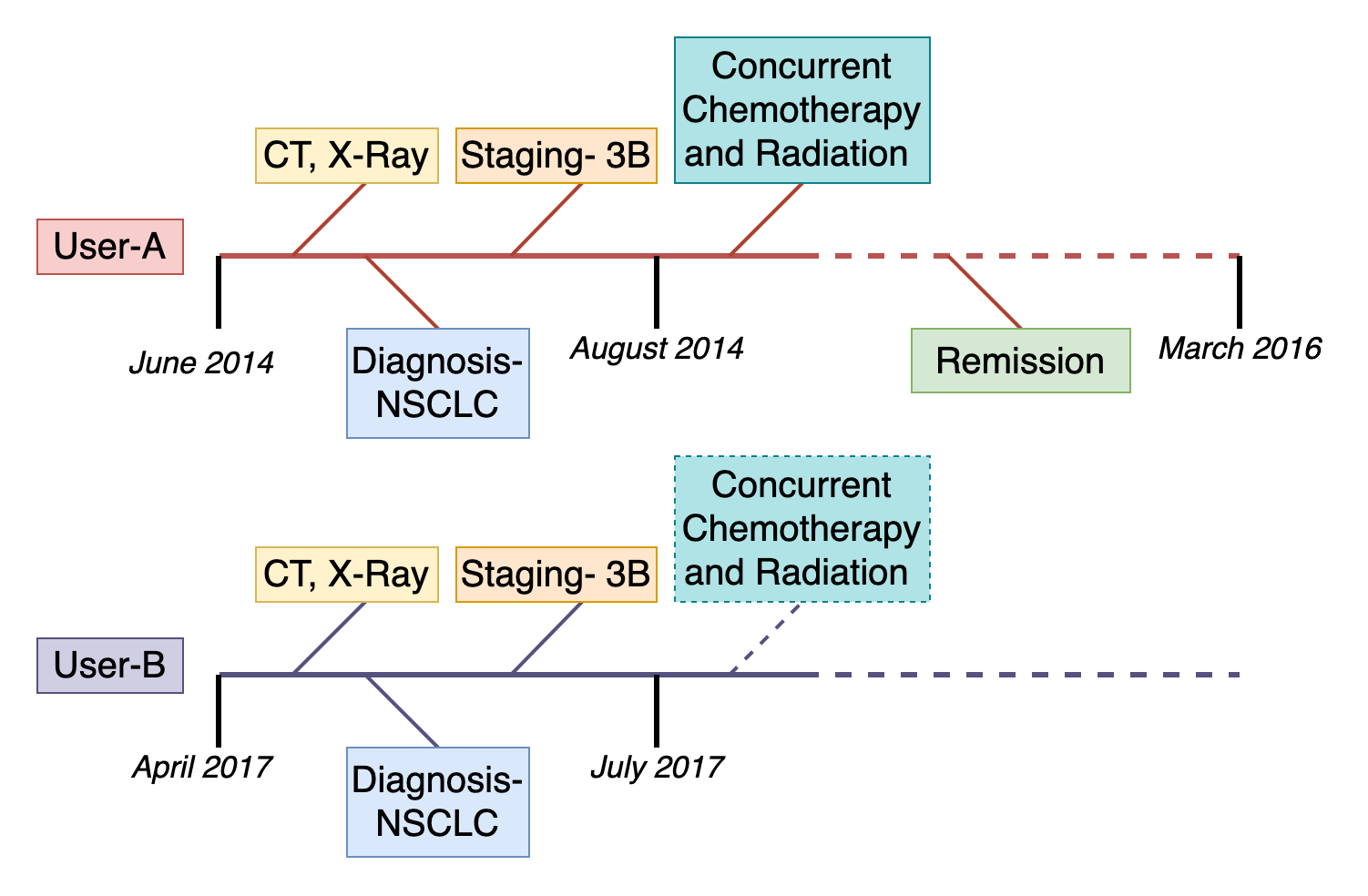}
    \caption{Similarity in relative disease timelines can help cluster similar users.}
    \label{fig:timalina}
\end{figure}
\begin{table*}
\caption{Differentiating existing models and our approach }

\center
\resizebox{\linewidth}{!}
{
\begin{tabular}{|c|c|c|c|}

\hline
\#& & Traditional Methods & Our Method\\
\hline
1 & Tag Prediction & Picking Tags from collection of tags \cite{wang-etal-2019-microblog} & Generating new words from vocabulary\\ 
2 & Tag Prediction Training & Increase probability of picked tags \cite{yuan2019attention} & Synthesize auxiliary sentences\\
3 & Objective & Prediction of tags only for input document \cite{shu2018content}\cite{wang2018dkn} & Prediction of tags for unseen future document\\
4 & User Clustering & Network interaction based \cite{klavsnja2018enhancing} & Based on disease progression similarity\\
5 & Vulnerability & Vulnerable to skewed tag distribution \cite{tuarob2013automatic}& Robust to tag distribution\\
6 & Evaluation Metrics & Precision@K (P@K), Recall@K (R@K) \cite{jiang2017user} & Cosine similarity in feature space along with P@K, R@K\\
\hline

\end{tabular}
}
\label{table:intro}
\end{table*}
\subsection{Research Question}
This study is designed to answer the following research question: Can the user profiles together with the traces of interactions of the user with the community (e.g., keywords used in past queries) and information from similar users be used to effectively predict contextually relevant  topic tags associated with the \textbf{{\em future}} information needs of users?

We hypothesize that patients with similar histories of disease progression or with similar courses of treatment would have similar information needs at comparable stages. In treating many diseases, e.g., cancer physicians follow a particular course of treatment that is customized based on the patient's profile and how they are responding to treatment. For example, the National Comprehensive Cancer Network (NCCN) has laid out guidelines for the treatment of non-small cell lung cancer \cite{ettinger2019nccn}. If two users, say $A$ and $B$, have similar histories of disease progression and treatment, we can expect that their information needs at comparable stages are likely to be similar. In \textbf{Figure \ref{fig:timalina}}, we illustrate an example with two users who have similar histories of disease progression and treatment. In this case, the topic recommendations for user $B$ can be improved by utilizing user $A$'s past keywords. In this case, user $B$ can be recommended articles related to \textit{``concurrent chemotherapy and radiation"}.

Based on the preceding discussion, we arrive at the following problem specification. For any given user $U$ in the community, with a user profile $B^U$, temporally ordered sequence of $n$ prior posts $\{P^U_{t-n}, \cdots, P^U_{t}\}$,  encoded by the corresponding keyword sets  $\{G^U_{t-n}, \cdots, G^U_{t}\}$, and indexed by their position in the sequence as well as the past information traces of all the similar users in the network, the task is to predict the topic tags  $G^U_{t+1}$ associated with a  hypothetical future post $P^U_{t+1}$ that describes the future information needs of the  user $U$. 


\textbf{\textit{Disambiguation:}}
``Tags" and ``Keywords" can be used interchangeably in our work as they both describe the abstract token used to represent a larger body of text. However, we try to differentiate by using ``Keywords" for tokens that have occurred prior to prediction and ``Tags" as tokens that are yet to be predicted (future).

\subsection{Contributions}

We propose an innovative approach for predicting future information interests of participants in online communities as they move through different stages in their disease or treatment. We report results of our experiments on an expert curated data set from  which we demonstrate the superiority of the proposed approach over the state of the art baselines with respect to accurate and timely prediction of topic tags (and hence information sources of interest). 

The key technical contributions of the work include:
\begin{itemize}
    \item A novel approach to predicting keywords associated with the future information needs of users (patients, caregivers) in an online health community. 
    \item A novel approach to collaborative-hybrid filtering based recommendation leveraging similarity of user disease progression timelines.
    \item A semantic evaluation of predicted tags along with the traditional metrics such as precision and recall.
\end{itemize}


\section{Data}\label{datasec}
We obtain our data from HealthUnlocked.com (HU), which is one of the largest online health communities. Users of the platform interact with other users through an easy-to-use interface as they seek information or emotional support from their peers   while managing their personal healthcare. A longitudinal study of HealthUnlocked has shown that it has positively impacted many patients \cite{barrett2016creating} in terms of health outcomes and engagement in care. The website has several sub-communities organized around specific health conditions or diseases. In our work, we focus on three such communities designated for lung cancer patients and their caregivers: \textit{Lung Cancer Support} , \textit{The Roy Castle Lung Cancer Foundation}  and \textit{British Lung Foundation}. Our choice of the communities was motivated by three considerations. First, managing chronic conditions and treatment regimes associated with lung cancer presents information needs that can be met through sustained interactions with the community. Second, the progression of disease and treatment for chronic conditions impacts the information needs of users over time, which poses challenges that are not adequately addressed by existing approaches in recommendation systems, personal health libraries, etc. Lastly, these communities represent  the largest fraction of the users of the online community.

Each user has an opportunity to create a profile which includes demographic data, e.g., age, gender, etc. as well as the health condition. Once registered, a user then may pick one or more communities to join. Once in a community, users can post a ``Main-Post" or post a ``Comment" on an existing ``Main-Post". Additionally, each ``Main-Post" can have associated ``keywords" at the time of posting. These user supplied keywords can be considered to be indicative of the information needs of the user at the time and act as ground truth for training and evaluating the proposed tag prediction system. 

We acquired  de-identified data from the lung cancer communities for the period between 2011 and 2020. From the three lung cancer communities, we have a total of 1711 unique users. Note that ethical (or IRB) approval was not needed for our study because the data were de-identified at source (at HealthUnlocked), there is neither direct interaction nor intervention with human subjects, and there is no means to link the de-identified data to specific subjects. For this study, we focus on the subset of anonymized users with at least one post and only valid posts with a minimum of three words or more are considered. The resulting data now filters down to 1032 unique users. Each user has an anonymized hashed User-ID that links their profile-bio, posts, and keywords to the user. Among the 1032 unique users, the average posts per user is about 2.4 (minimum is 1 and maximum is 68). Each post has on average  3.2 keywords (minimum is 2 and maximum is 10). About 55\% (574) of the users have a profile-bio. An example of a user's information is provided in \textbf{Figure \ref{fig:inpsnippet}}.

Due to the unstructured nature of the text in our data set, identifying users whose information traces and temporal pattern of disease and treatment progression may be predictive of the future information needs of a target user presents significant challenges.  To address this need, we restructured the user data as described below.

  \begin{figure}[h]
    \centering
    \includegraphics[width=1\columnwidth,height=0.23\textheight]{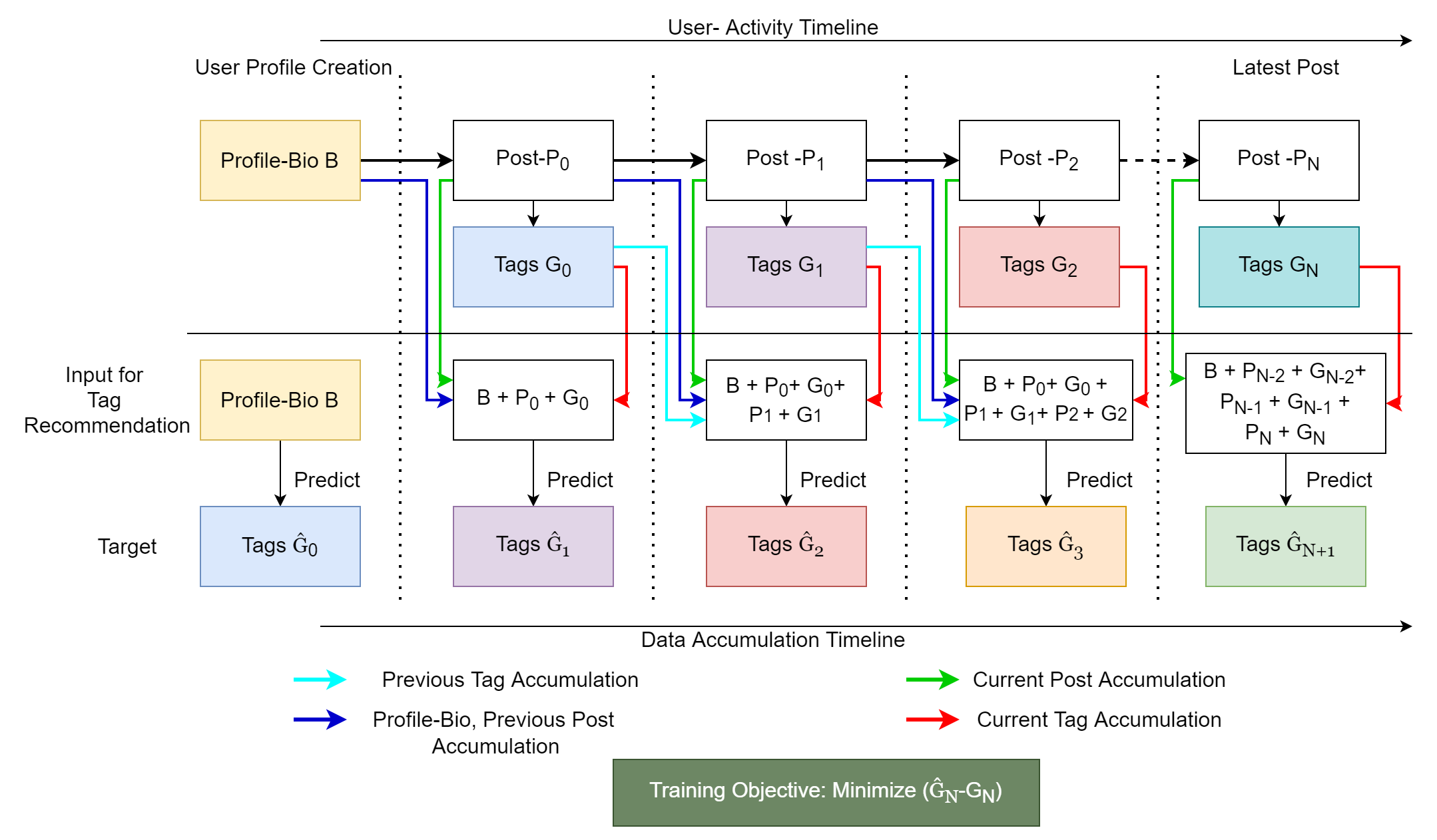}
    \caption{User Timeline and Data accumulation for temporal Tag Prediction}
    \label{fig:my_label3}
\end{figure}
\subsection{Structuring the information traces of users}
We organize the information associated with each user, including the profile bio and the traces of the user's interaction with the community (information seeking posts or comments) as shown in \textbf{Figure \ref{fig:my_label3}}. Here we assume that a user $U$ creates a profile-bio $B$ before they post their first post $P_0$. At time $T_{-1}$, we use the bio to predict tags $\hat{G}_0$ where $\wedge$ indicates the prediction. This prediction is made before the first post has been created. As time progresses, the user posts $P_0$ at time $T_0$ with keywords $G_0$, We then accumulate the profile bio $B$, the post $P_0$ and keywords $G_0$ to predict the tags $\hat{G}_1$ of a future unseen post $P_1$. We continue to do this with a temporal window of $Pn$ posts in the past along with the current post to predict tags of a future post. At time $T_n$, though we have $n$ posts from the past, using all the posts and their keywords could add too much noise into the topic tag prediction system. We empirically question the influence of history (see \textbf{Section \ref{infhist}}) and determine the ideal $Pn$. 
We then use only the last $Pn$ posts, their associated keywords, and the profile bio which remains unchanged throughout time $T_0$ to $T_n$, as well as the current post and corresponding keywords to predict a future topic tag $\hat{G}_{n+1}$. 

\subsection{Example Generation}
\begin{figure}[h]
    \centering
    \includegraphics[scale=0.149]{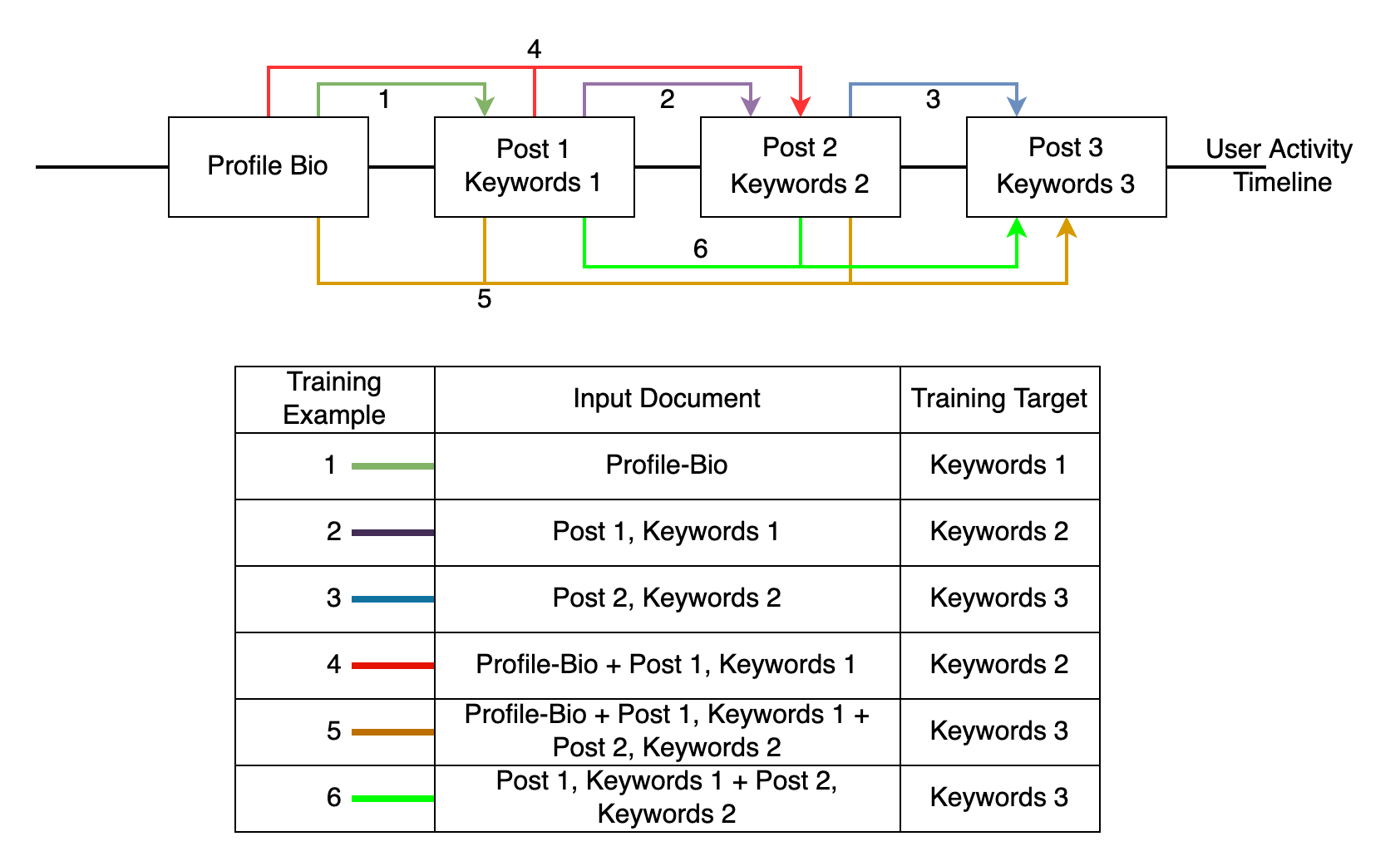}
    \caption{A timeline from a single user with a profile-bio and 3 posts is converted into 6 training examples.}
    \label{fig:trainexample}
\end{figure}

To create a training example, we need an input sequence (input text document) containing profile-bio, posts and keywords and a target sequence (Tags of future post). This is commonly known as sequence-to-sequence training. In our setting, 
we increase our training example pairs by extracting all possible pairs of data as shown in \textbf{Figure \ref{fig:trainexample}}. In order to maintain chronological patterns in the data, we pick pairs that have information from consecutive time steps. For example, if \textit{Post-1} and \textit{Keywords-1} are from the latest time step, the target can only be \textit{Keywords-2}. This input set cannot be paired with \textit{Keywords-3} as \textit{Post-2} and \textit{Keywords-2} are not included the input. Using such a rule, a single user's timeline is converted into six chronological training pairs as shown in \textbf{Figure \ref{fig:trainexample}}. Following this rule, we obtain a total of 10,791 training pairs from the 1032 unique users. From this dataset, we use 8415 random pairs (78\%) for training and the remaining 2376 (22\%) pairs for testing.

\subsection{Tag Statistics} \label{tagstat}

\begin{figure}[h]
    \centering
    \includegraphics[width=0.48\textwidth,height=0.3\textheight]{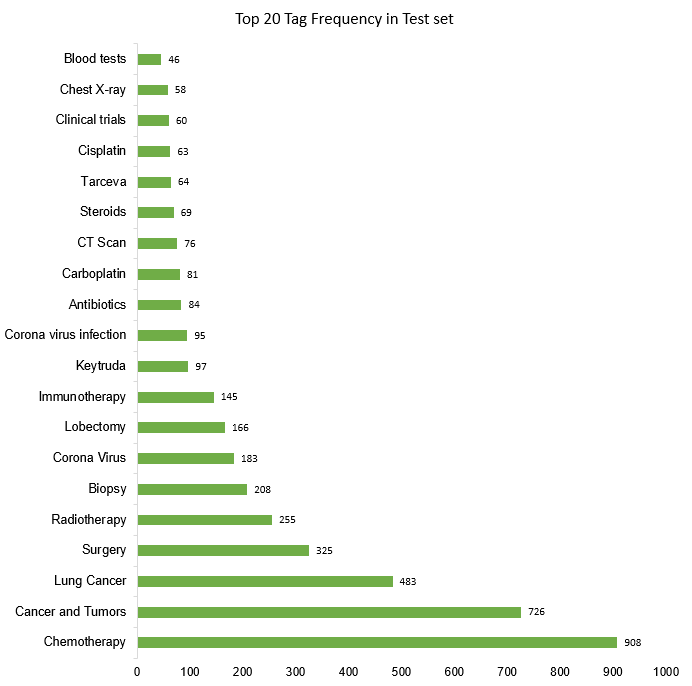}
    \caption{A plot of 20 most frequently occurring tags in the test set}
    \label{fig:hist}
    
\end{figure}

Tag prediction can be approached through a tag-classification task. However, if the dataset has uneven tag distribution like ours, the model is more likely to fail due to high bias towards the more frequent tags. Additionally, with a large number of tags, classification based models tend to perform poorly. Our training set has 472 unique tags while the test set has 375 tags. Among the 375 tags, 18 of them are not in the training set and are completely new words. However, since these words occur in the posts of the users, they are a part of the vocabulary that the models can observe and thus are not deleted from the test set. \textbf{Figure \ref{fig:hist}} shows the distribution of the top 20 tags in the test set. It is evident that \textit{`Chemotherapy'}, \textit{`Cancer and Tumors'} occur several times more frequently than other tags. Tag prediction models designed around classification could incorrectly pick frequently occurring tags leading to poorer performance. Additionally, their metrics tend to be sensitive to the tag frequency in the training data. We address this issue of skewness and the resulting bias in \textbf{Section \ref{dbt}}.

\section{Methodology} \label{methodo}

Recent advancements in the field of natural language processing (NLP) have opened a realm of possibilities due to a better understanding of context from text data.  Bidirectional Encoder Representations from Transformers \cite{devlin2018bert} have further allowed text representations to be learnt from vast sources like the Book Corpus and the English Wikipedia with more than 3300 million words. Their utility has been recognized in the bio-informatics community through the proposal of Bio-BERT \cite{lee2020biobert}, an extension of the BERT, trained over the PubMed and PMC medical word corpora. The extension of the pre-trained models helps in inferring context,  more effectively than the traditional frequency based models, from text containing medically relevant words including those from online health communities. We leverage the Bio-BERT to extract illness-relevant information and predict future topics of interest based upon the users' needs.

Our pipeline is intended to extract context from the unstructured data and recommend accurate and time-relevant tags for each user in the online health community database. We make several modifications from works in the literature to accommodate our data and increase the overall accuracy of tag prediction. Among these, is our usage of Bio-BERT model as a sentence generator instead of a simple bag-of-words classifier. Additionally, we show the need for collaborative filtering but move away from traditional K-means method to a disease timeline based clustering method. In this section, we will describe our proposed tag prediction procedure, the model architecture, the objective function for performing the training and the metrics used to compare the models in our experiments.

\subsection{Tag Prediction}
\label{taggen}
\begin{figure*}[h]
    \centering
    \includegraphics[width=0.8\textwidth,height=0.36\textheight]{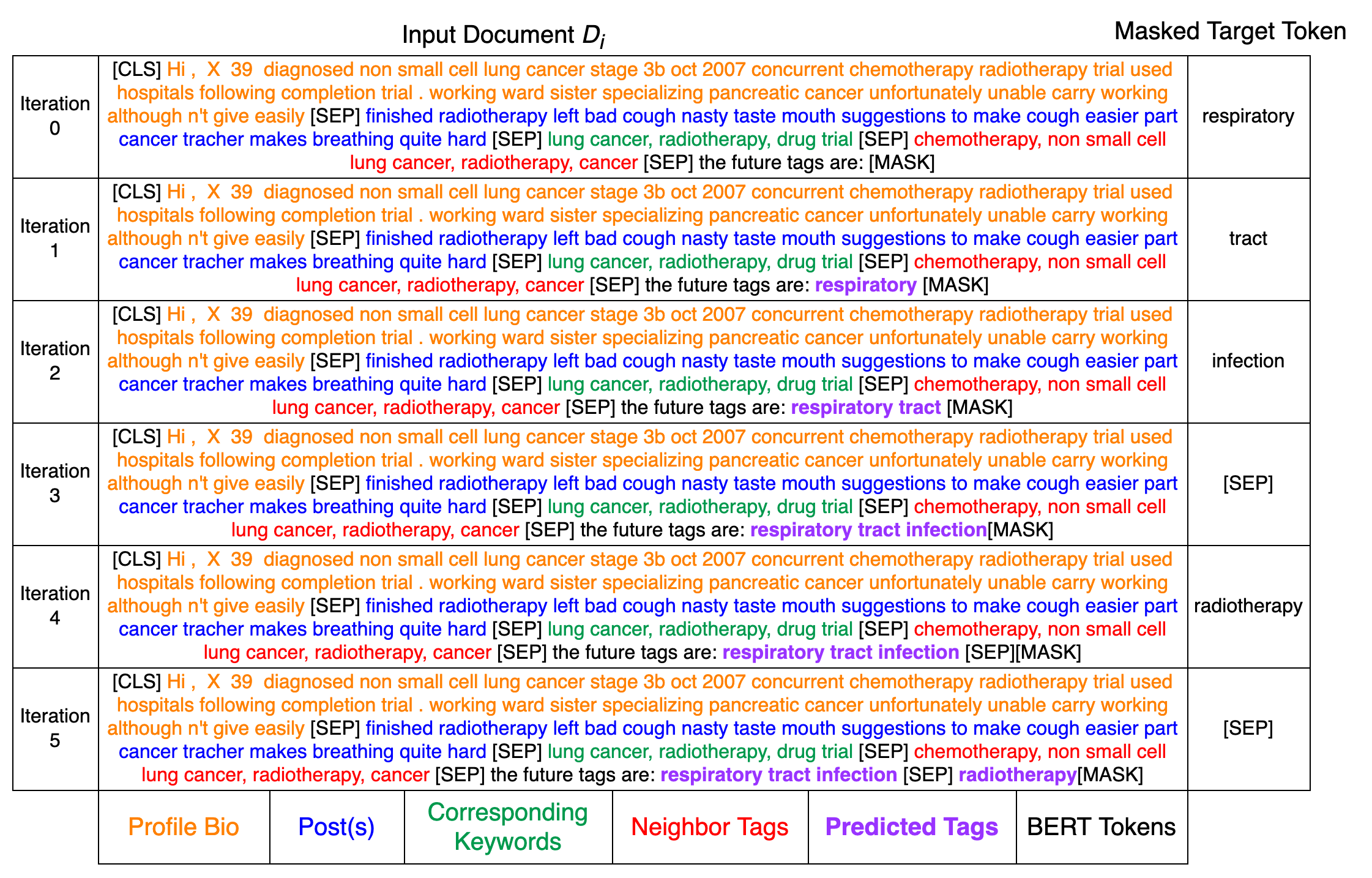}
    \caption{Example of a sequential training routine of $BBERT _{Tg}$ with 2 tag generations. The input document has multiple sentences and each sentence is separated by $[SEP]$ token. Each iteration has a specific target.  The target token at each iteration is appended at the position of the $[MASK]$ token in the next iteration. A $[SEP]$ token is added after each tag is predicted.}
    \label{fig:trainseq}
\end{figure*}

Our primary objective in this work is to predict future topic tags for users in order to forecast their medical interests. Unlike existing tag prediction algorithms, we do not conform to a classification approach where a model picks the most probable tag from a set of tags. Instead, we perform an auxiliary sentence generation (ASG) task. The ASG task is similar to sentence completion where the last word is masked out using the $[MASK]$ token. In our set-up, the sentence generated is a set of comma separated future tags with no grammar or punctuation. We fine-tune a pretrained Bio-BERT model to read a given input document and pick a word from its entire vocabulary that fits the context and position of the mask. A similar training approach has been employed by \cite{chan2019recurrent} for generating questions based on a paragraph of context and an answer sentence. We, however, predict comma separated words which are actually the future tags. Since we are using a Bio-BERT model Tag generation, we refer to our model as \textbf{BBERT$_{Tg}$}. An example of the training process has been shown in \textbf{Figure \ref{fig:trainseq}}. First, we have a $[CLS]$ token to indicate the start of the input document. The text in Orange is the profile-bio followed by a $[SEP]$ separation token. Similarly, in Blue, we have the past and current posts, the Green text corresponds to tags from previous and current posts, the tags from neighboring users or shared tags (see \textbf{Section \ref{modimprov}})  are in Red and the predicted future tags are in Purple. We experimented with the positioning of these individual elements in the input document. However, they did not lead to a significant difference in the results. 
Having accumulated all the parts of the input document, we first clean the text through removal of stop words and lemmatization. Following this, the words are converted to fixed length vectors using a BERT word-embedding and tokenization module which converts the sentences into tokens based on the position of the words in the sentence and the position of the sentence in the input document. This encoded data is then fed to the Bio-BERT model along with the BERT-tokens like $[CLS]$, $[SEP]$,$[MASK]$ etc.

The pre-trained Bio-BERT model uses the originally prescribed BERT$_{base}$ model with 12 layers, 768 hidden dimensions and 12 attention heads. We followed the training procedure set by \cite{chan2019recurrent} with the usage of an Adamax optimizer set with an initial learning rate of 5e-5. The training is performed over 15 epochs with 8415 examples while 2376 examples are used for testing. We use softmax as the final layer of the model where the Bio-BERT computes the probabilities of 200K+ words. Following literature, we adapt the Beam search \cite{freitag2017beam} method to pick the top words and decode the predicted Bio-BERT output into regular words (tags).

\textbf{Modification for Bio-BERT with long text Documents}

The traditional transformer model is meant for short sentences as computing attention over large input documents is computationally expensive and requires lengthy training. Our input documents range from about 14 words to 550+ words when multiple data sources like profile-bio, posts, keywords etc. are combined.  We modify our fine-tuned model based on the recommendations provided by \cite{beltagy2020longformer} where a ``longformer" is proposed. This reduces the attention computation from quadratic to linear space. Gated Recurrent Units (GRUs) and Long-Short Term Memory (LSTM) models, though not entirely bidirectional in attention, are however capable of handling such large documents without any modifications.

In the next two subsections, we shall detail the training objective function and the metrics we use for evaluating the model.

\subsection{Objective Function}

In our sequence-to-sequence based learning approach the input sequence is the input document containing profile bio and posts and the output sequence is the target tags set. Both the sequences are assumed to come from a single long sentence where the target sequence is masked. We first tokenize the input sequence $x$ into $n$ tokens, $\{x_i\}, i=1,2,...n$, and predict the sentence fragment $x^{p:q}$ where $p$ and $q$ are positions such that $p$$<$$q$$<$$n$ using the specially masked input which can be denoted as $x^{/p:q}$. Unlike traditional tasks like translation, $x$ does not have a paired $y$ but we instead have to rely on the conditional probability of a token appearing next in the sequence given the past tokens (until the $r$-th position, $p \leq r \le q$ ) of the sequence. Let us denote the domain from which $x$ is obtained as $\chi$, the log likelihood objective to learn the model parameters $\theta$ then becomes:
\begin{equation}
L(\theta;\chi)=\frac{1}{| \chi |}\Sigma_{x \in \chi}\log{P(x^{p:q}|x^{/p:q}; \theta )} \label{eq0}
\end{equation}
Expanding the conditional probability as a product of probabilities at each position between $p$ and $q$, we obtain \textbf{Equation \ref{eq1}}. Here, $r$ is the position at which the $[MASK]$ token is placed. 
\begin{equation}
L(\theta;\chi)=\frac{1}{| \chi |}\Sigma_{x \in \chi}\log{\prod_{r=p}^{q}P(x_{r}^{p:q}|x_{<r}^{p:q},x^{/p:q}; \theta )} \label{eq1}
\end{equation}
The position, or $r$, can be controlled to dictate the amount of prior information in the input document. In our experiments, we always have a non-empty set of tokens for prior data coming from either Profile Bios or Posts. However, \textbf{Equation \ref{eq1}} can be used with null prior data where it will generalize into the OpenAI GPT model for generating text~\cite{song2019mass}.

\subsection{Metrics}
To evaluate the tag generation model, we can view it as a recommendation system which recommends topic tags. This is similar to medication recommendation evaluation by \cite{9756911}. We can generate any number of tags once the model is trained as this would only involve generating longer sentences. Traditionally, if $K$ is the number of tags recommended, Recall@$K$ and Precision@$K$ are used to evaluate the success of the recommendation.
In order to enable fair comparison using both Recall and Precision, we use the popular F1@K metric where K is the number of predictions made and F1 is calculated with equal weights to both recall and precision.

We go beyond the traditional evaluation and calculate cosine similarity in the feature space between the recommended tags and the ground truth. This is primarily because the tag generation network is not picking words from a list but instead generating new words with a certain degree of variety. To calculate the cosine similarity, we first extract the word embeddings (features) of the $K$ predicted tags and individually check their similarity to the word embeddings of the ground truth tags.  If the ground truth tag is \textit{`radiotherapy'} and the predicted tag is \textit{`radiation therapy'} the penalty will not be too large.


\section{Experiments}\label{expe}

\subsection{Individual User Model Ablation}
In the first set of experiments, collaborative filtering is not performed and each user is an independent data point. We determined the ideal settings for the tag generation model based on the following variables- \begin{itemize}
\item Accumulation of different elements of the input document- (1) Profile Bios only ($B^U$), (2) Profile Bios \& Posts ($B^U$+$P^U$) (3) Profile Bios , Posts \& Corresponding keywords ($B^U$+$P^U$+$G^U$).  
\item The number of past posts ($P^U_n$) to use for predicting future tags
\end{itemize}

\subsubsection{Input Document Accumulation}
To understand the importance of different elements in the data, we conduct three experiments with increasing amount of information presented in the input document. We start with the profile bio if available and then add posts and their corresponding keywords in the subsequent trials. For this experiment, we use 2 past posts along with the current post ($P_n$=3).
In the $B^U$ run (profile bio only), the model is expected to predict all future interests based on a fixed profile-bio which the user has provided during the creation of their account. In many cases, the user has interests unrelated to the profile-bio as they are not required to update this field regularly. However, the overarching context related to their condition is sometimes provided in the bios. Posts $P^U$ are comparatively more in tandem to the users' current interests. They describe what the user is actively querying and make predictions of the future tags easier. Keywords $G^U$ from the past and current posts additionally help the model understand abstraction between the post and their corresponding keywords. The average results with 4 runs each are presented in \textbf{Table \ref{tab:accu}}.


\subsubsection{Influence of History} \label{infhist}
Our dataset is primarily based on chronic health conditions, therefore, utilizing the past information through past posts is vital for predicting future topics of interest. However, for users with many posts, accumulating all the posts may cause the network to pay attention to noisy topics occurring in the past. Additionally, extremely large documents have larger computational costs in terms of feature extraction. For this, we determine the ideal number of previous posts ($P^U_n$) required. $P^U_n$=1 corresponds to using only the current post while $P^U_n$=4 accumulates the last 3 posts along with the current post. For this experiment, we use the $B^U+P^U+G^U$ (referred to as full mode) of data accumulation. \textbf{Table \ref{tab:hist}} details the average results for 4 runs of this experiment
\begin{table*}

\caption{Influence of Data Accumulation in Input document, $P_n$=3, Averaged over 4 individual runs}
\label{tab:accu}       
\center
\resizebox{\linewidth}{!}
{
\begin{tabular}{|c| c | c| c | c | c | c|c|c|c|c|c|c| }

\hline
 & \multicolumn{4}{|c|}{$K$=1} & \multicolumn{4}{|c|}{$K$=3} & \multicolumn{4}{|c|}{$K$=5} \\
\hline
Input Document & Recall & Precision & Cosine Similarity & F1 & Recall & Precision &  Cosine Similarity &  F1 & Recall & Precision &  Cosine Similarity & F1\\
\hline
$B^U$ (Only profile bio) & 0.205 & 0.487 & 0.491 & 0.288 & 0.430 & 0.296 & 0.476 &0.350 & 0.483 & 0.187 & 0.428& 0.269\\
$B^U$+$P^U$ & 0.274 & 0.791 & 0.866 & 0.407  & 0.517 & 0.442 & 0.830& 0.476 & 0.594 & 0.363 & 0.826 & 0.450\\
$B^U$+$P^U$+$G^U$ (Full mode) & \textbf{0.296} & \textbf{0.874} & \textbf{0.909} &
\textbf{0.442}&
\textbf{0.590} & \textbf{0.584} & \textbf{0.891} &
\textbf{0.586}&
\textbf{0.681} & \textbf{0.402} & \textbf{0.887}&
\textbf{0.505}\\
\hline

\end{tabular}
}
\end{table*}
\begin{table*}

\caption{Influence of Past Posts in Input document with full mode data accumulation, Averaged over 4 individual runs }
\label{tab:hist}       
\center
\resizebox{\linewidth}{!}
{
\begin{tabular}{|c| c | c| c | c | c | c|c|c|c|c|c|c|c }

\hline
 Number of past posts in & \multicolumn{4}{|c|}{$K$=1} & \multicolumn{4}{|c|}{$K$=3} & \multicolumn{4}{|c|}{$K$=5} \\
\hline
Input Document & Recall & Precision & Cosine Similarity & F1 &  Recall & Precision & Cosine Similarity & F1  &  Recall & Precision & Cosine Similarity & F1\\
\hline
$P_n$=1(Only current post) & 0.271 & 0.791 & 0.866 & 0.403 & 0.536 & 0.453 & 0.856 &0.491 & 0.597 & 0.381 & 0.832 &0.465\\
$P_n$=2 (One previous post) & 0.277 & 0.820 & \textbf{0.909}&0.414 & 0.545 & 0.567 & 0.879 & 0.555 & 0.657 & 0.395 & 0.876 & 0.493\\
$P_n$=3 (Two previous posts)  & \textbf{0.296} & \textbf{0.874} & \textbf{0.909} & \textbf{0.442} & 0.590 & \textbf{0.584} & \textbf{0.891}& \textbf{0.586} & \textbf{0.681} & \textbf{0.402} & \textbf{0.887} & \textbf{0.505}\\
$P_n$=4 (Three previous posts)& 0.286 & 0.856 & 0.887 &0.428 & \textbf{0.593} & 0.577 & 0.887 & 0.584& 0.671 & 0.381 & 0.852&0.486\\
$P_n$=5 (Four previous posts) & 0.284 & 0.823 & 0.871 & 0.422 & 0.555 & 0.502 & 0.874 & 0.527 & 0.649 & 0.377 & 0.834 &0.476\\
\hline

\end{tabular}
}
\end{table*}

\subsection{Model Improvement Through Collaborative Filtering } \label{modimprov}
Our approach for model improvement comes from the intuition that \textit{``similar users seek similar information"}. Collaborative filtering is an approach to view a user as a parametric function of interests defined by other similar users in the community. Users within a group or a cluster tend to seek similar information or products. Multiple recommendation systems including \cite{kaur2018efficient} and \cite{tran2018hashtag} utilize collaborative filtering to extract a deeper understanding of user needs and provide recommendations based on other users in the group. 

Identifying similar users is a crucial step in performing collaborative filtering. In our setting, with healthcare data, we identify similar users by matching relative disease timelines. Users with similar illness progression tend to search for similar information. Information search history of users who have already traversed the timeline can be utilized to recommend topics to users trailing along similar paths. We hypothesize that the similarity in treatment profiles for chronic conditions like non-small cell lung cancer (NSCLC), can be leveraged to recommend information at different time intervals based on the disease progression. 

\textbf{Neighbor Tags:}
For a query user $U$ with a post $P^U$ and keywords $G^U$ posted on date $T$, we identify nearest neighbors $h_1,h_2,..h_n$ using one of the three methods described below. For each neighbor $h_i$, we check for posts that have occurred prior to $T$. Then the keywords associated with such a post will be referred to as \textit{Neighbor Tags} and are appended to the input document after the posts and their corresponding keywords (see \textbf{Figure \ref{fig:trainseq}}). The entire input document is then used for future topic tag prediction.

In order to perform user similarity matching and collaborative filtering, we experimented with three approaches. 
\begin{itemize}
    \item Grouping of user profile-bios using K-means clustering.
    \item Using profile-bio text features extracted from Bio-BERT model for feature space clustering (t-SNE).
    \item Novel disease timeline based similarity matching of users.
\end{itemize}

\subsubsection{K-Means Clustering}
K-means is a well established method for performing clustering of users in a collaborative filtering setting for the purpose of recommendation \cite{zarzour2018new}.
In this baseline approach, we first performed primary cleaning through stop word removal and lemmatization. We then vectorized the words using scikit-learn \cite{scikit-learn} \textbf{TfidfVectorizer} package and used a gap statistic based elbow approach \cite{tibshirani2001estimating} to identify the ideal number of clusters. We experimented with 2 to 80 clusters  and found the characteristic elbow at 34 clusters. Following this, we fit the profile bios into 34 multi-user clusters.  The average number of users in each of these clusters were approximately 11. For a query user $Q$, we randomly pick 3 users from the same cluster that $Q$ belongs to. When 3 users were not available in a given cluster, we dropped the query user and moved to the next. To further avoid sampling bias, we repeated this experiment 4 times to randomly sample neighbors in a cluster. We only selected those neighbor tags which had occurred prior to $Q$'s current post. We noticed an overall lower (compared to no neighbor tags) tag-recommendation F1 score when neighbor tags provided by K-means were included in the BBERT$_{Tg}$ input document. This has been detailed in \textbf{Table \ref{tab:newone}} (row 3).

\subsubsection{Bio-BERT Feature Clustering}
A major drawback of the K-means approach is the number of fixed clusters. In an online health community setting, users may be close to multiple clusters. Additionally, the embeddings extracted from profile bios may not sufficiently represent similarity (or dissimilarity) in the feature space due to the simplicity of the encoder used. To overcome these issues, in our second approach, we replaced the \textbf{TfidfVectorizer} feature extractor with a Bio-BERT model in order to improve the semantic representation of profile-bios in the feature space. We reduced the BERT feature vectors to a two dimensional t-Stochastic Neighborhood Embedding (t-SNE) space and found the 3 nearest neighbors ($h$) using simple euclidean distances as a measure of similarity. The absolute location of points on a t-SNE plot is not an indication of quantitative similarity. However, we can use the relative euclidean distance between two or more points averaged over 4 independent t-SNE runs to provide a closer estimate of the nearest neighbors. In this approach, for a given query user, $Q$, the number of neighbors is not limited (soft clustering) and any number of neighbors can be extracted based on distance from query user. The keywords of the neighbors occurring prior to current post were then used as neighbor tags in the input document. Similar to the K-means baseline approach, we evaluated this clustering method through changes in the tag-recommendation F1 score. This has been recorded in \textbf{Table \ref{tab:newone}} (row 4). 

\begin{figure*}
    \centering
    \includegraphics[scale=0.1]{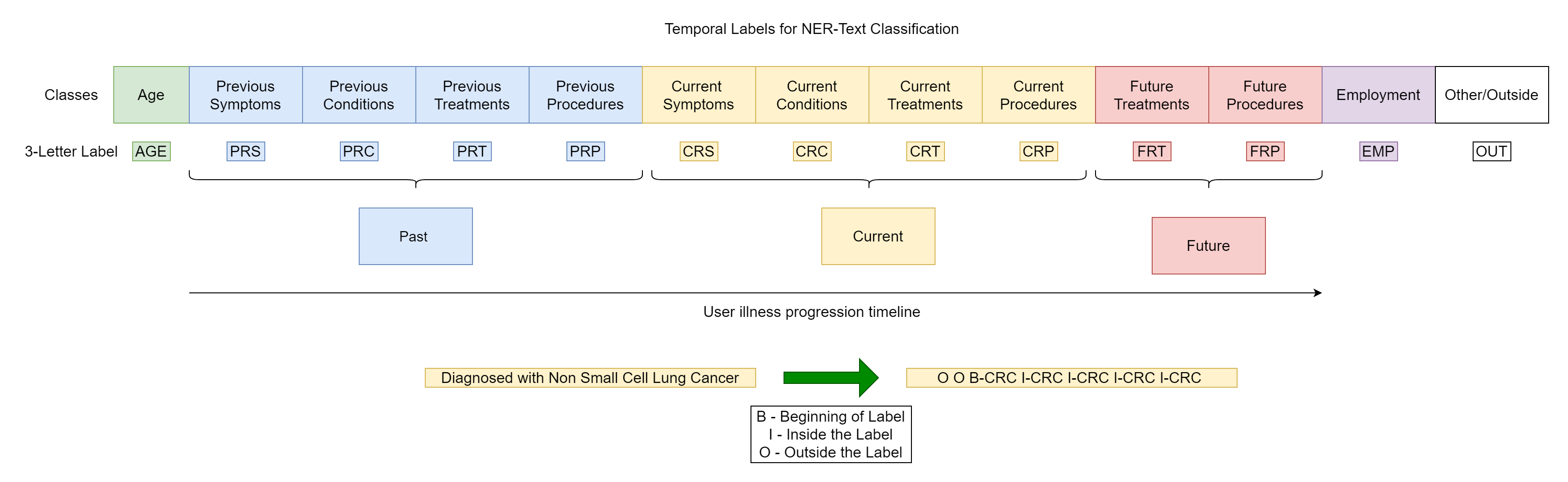}
    \caption{Temporal Named Entity Recognition task Labels based on illness progression}
    \label{fig:my_label4}
\end{figure*}
\subsubsection{Disease Timeline Matching}
In this novel approach, we implemented the following steps in order to extract neighborhood clusters and share tags within them.

\begin{itemize}
    \item The unstructured profile-bios were converted into a structured 13-column disease timeline using a Named Entity Recognition (NER) BERT model 
    \item A Bio-BERT model was used to extract features from the disease timelines and were clustered in the feature space using t-Stochastic Neighborhood Embedding
    \item The $h$ nearest neighbors are identified using euclidean distance in the two dimensional t-SNE feature space.
    \item  Keywords of neighbors which have occurred prior to query post are used as neighbor tags ($St$). 
    \item Neighbor tags are then appended to the input document before being provided to the BBERT$_{Tg}$ model. 

\end{itemize}
\underline{\textbf{Conversion of Profile-Bios:}}
First, we converted the unstructured profile bios to a structured temporal timeline. In order to provide structure to the profile bios, we contextually sorted the text into various classes. The classes created are shown in  \textbf{Figure \ref{fig:my_label4}}. Where \textit{``Other"} class corresponds to words that do not belong to any of the other 12 classes.
``Current" refers to the time when the bio was written, ``Past" and ``Future" are with respect to the ``current" time. As it can be observed, the data is forced to take a temporal shape in order to extract user-disease timeline. This annotation was done by two subject experts who manually read each profile bio and filled in the corresponding columns. The annotators read the entire profile-bio and then determined if parts of the text can be put into one of the 13 classes mentioned above. 
To verify the usability of the labeled dataset, we compared the similarity across the two annotators and obtained a cosine similarity of 0.811 and a Cohen's kappa of 0.784. According to \cite{mchugh2012interrater} our score for inter-rater reliability and agreement was ``substantial" and 0.16 short of ``perfect agreement". We observed the lowest agreement in the \textit{Employment} class. An intersection of the labels between the annotators is used as ground truth for the NER text classification task.

\begin{figure}[h]
    \centering
    \includegraphics[scale=0.33]{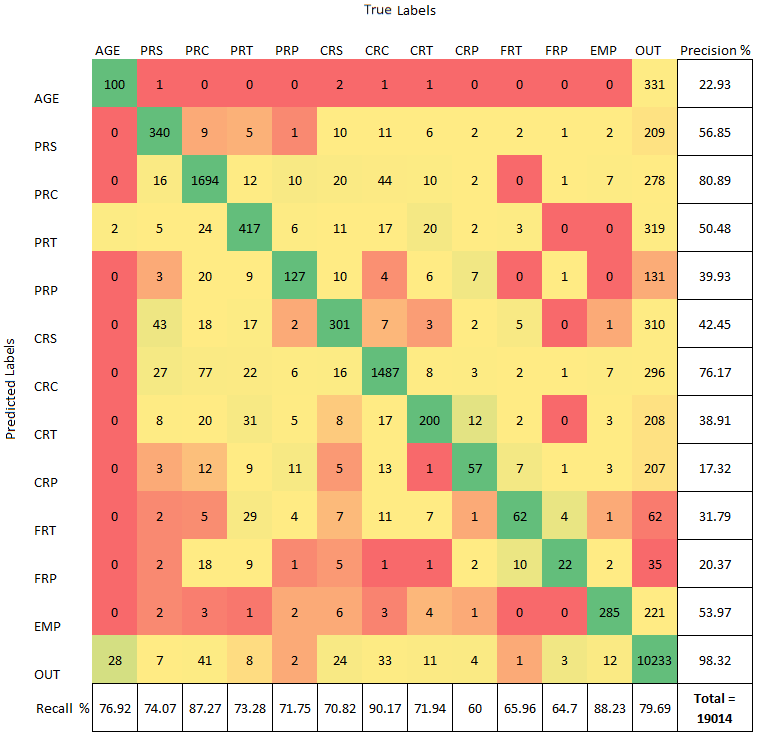}
    \caption{Confusion Matrix from Bio-BERT NER classification of profile-bio text in test-set}
    \label{fig:my_label6}
\end{figure}

In order to automate this conversion between unstructured profile bios to structured timeline data, we trained a NER model to perform a 13-class classification of profile-bio text. We perform classification only to obtain the temporal disease timeline and not the topic tags. The classes correspond to the individual columns in the temporal timeline of \textbf{Figure \ref{fig:my_label4}}. We trained two models namely (i) Bio-BERT\cite{lee2020biobert} and (ii) LUKE-Deep Contextualized Entity Representations with Entity-aware Self-attention \cite{yamada2020luke}. The Bio-BERT is only fine-tuned on our dataset while LUKE was trained entirely from scratch on our dataset. This was done to evaluate the value of information provided by the Bio-BERT pre-training. For both models, we used the B,I,O labelling procedure (Beginning of a label, Inside a label and Outside a label) to maintain standardized practices followed in the NLP community and to increase reproducibility. In the training set, we used 414 profile bios with 44,320 total words. In the non overlapping test set, there were 19,014 words from 160 bios. The models were trained for 25 epochs on a Nvidia DGX Tesla V-100 GPU for approximately 3.5 hours for fine-tuning Bio-BERT and 18 hours for training the LUKE. The Bio-BERT model outperformed the LUKE classifier in the following metrics- (a) The overall accuracy of the Bio-BERT model was 80.59\% while LUKE achieved 73.63\%. (b) The average recall percentage for Bio-BERT was 74.98\% while only 66.29\% for LUKE. The precision for each of the classes is low due to the large number of words in the ``Other (OUT)" category. The imbalance in the dataset skews the predictions which lead to a high false positive rate (see \textbf{Figure \ref{fig:my_label6}}). The confusion matrix for Bio-BERT along with its class-wise precision and recall is shown as well. For all the downstream tasks that follow, we use the Bio-BERT model fine tuned on our dataset as the primary feature extractor.

\begin{figure}[h]
    \centering
    \includegraphics[scale=0.135]{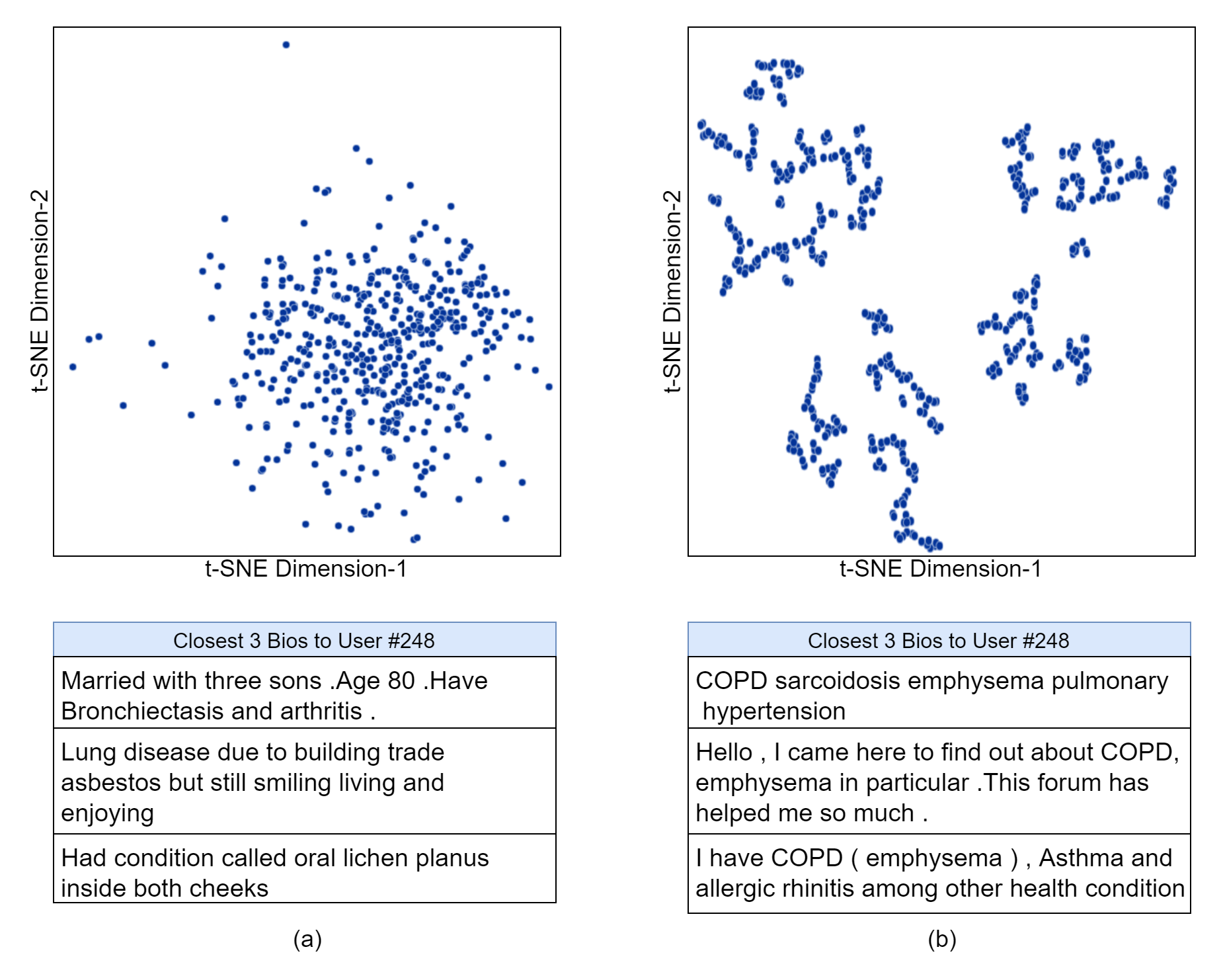}
    \caption{(a) t-SNE plot of features extracted directly from profile-bios of 414 users, Closest 3 bios (shortened for image) to random user \#248; (b) t-SNE plot of features extracted from NER temporal-classified user timeline of 414 users, Closest 3 bios (shortened for image) to random user \#248}
    \label{fig:my_label5}
\end{figure}

\underline{\textbf{Feature Extraction and t-SNE:}} The next step in the clustering process is to perform unsupervised clustering based on similarities in the column values. The 13 columns can be treated as a sentence vector for feature extraction. We employed the previously used Bio-BERT fine tuned on profile-bios to extract the features of the structured text data and map it to a two dimensional t-distributed stochastic neighbor embedding (t-SNE). Using the plot, we infer soft-clusters\cite{kim2020patent} based on user grouping in the 2-D space. Features extracted from profile bios directly and features from the temporal timeline are shown in the t-SNE plots acquired after 750 iterations in \textbf{Figure \ref{fig:my_label5}}. As it can be seen, features extracted from temporal timelines following NER form natural clusters compared to features directly extracted from profile bios. This is primarily due to the noise removal inherently performed by the NER task. For illustration, we pick a random user (\#248) and check the nearest 3 users' profile-bios. Inferred from \textbf {Figure \ref{fig:my_label5}b}, the clustering is around \textit{COPD} and \textit{Emphysema}.

For the purpose of clustering, we find the nearest \textit{h} users to the primary user using Euclidean distances in the t-SNE feature space similar to the previous approach. We accumulate all  keywords of the neighbourhood users, occurring prior to current post and append it to the input document of the primary user. The impact of neighbor tags from 3 neighbors clustered with with disease timeline similarity is shown in \textbf{Table \ref{tab:newone}} (row 5).

Having established the ideal clustering model, we experiment with number of neighbors to determine influence of similar users in the community. \textbf{Table \ref{tab:neigh}} illustrates the effect of collaborative filtering, where, row 1 has the original full mode model and the subsequent rows present the improved model (BBERT$_{Tg}$) with neighbor tags. 

At the end of clustering, a particular user has the following items in their input document: (a) The profile-bio, (b) The previous posts and their corresponding keywords, (c) The current post and its keywords and (d) Neighbor tags from  similar users. This input document is used to predict future tags $G^U_{T+1}$.
\begin{table*}

\caption{Comparison of user grouping approaches with $h$=3 neighbors, $Pn$=3 and full mode data accumulation, Averaged over 4 individual runs; No neighbors and random neighbors are included as baselines }

\label{tab:newone}       
\center
\resizebox{\linewidth}{!}
{
\begin{tabular}{|c| c | c| c | c | c | c|c|c|c|c|c|c| }

\hline
 & \multicolumn{4}{|c|}{$K$=1} & \multicolumn{4}{|c|}{$K$=3} & \multicolumn{4}{|c|}{$K$=5} \\
\hline
Clustering Method & Recall & Precision & Cosine Similarity & F1 &  Recall & Precision &  Cosine Similarity & F1 &  Recall & Precision &  Cosine Similarity & F1\\
\hline
No Clustering, No Neighbors & 0.296 & 0.874 & 0.909 & 0.442 & 0.590 & 0.584 & 0.891 & 0.586 & 0.681 & 0.402 & 0.887 & 0.505\\
No Clustering, 3 Random Neighbors & 0.277 & 0.813 & 0.763 & 0.413 & 0.510 & 0.526 & 0.730 & 0.517 & 0.597 & 0.383 & 0.700 & 0.466\\
K-Means of Profile-Bios \cite{zarzour2018new} & 0.290 & 0.865 & 0.812 & 0.434 & 0.572 & 0.545 & 0.742 & 0.558 & 0.610 & 0.384 & 0.765 & 0.471\\
BBERT Features of Profile-Bios & 0.301 & 0.880 & 0.911 &0.448 & 0.629 & 0.612 & \textbf{0.892} & 0.620 & 0.708 & 0.491 & 0.887 & 0.579\\
BBERT Features of NER Disease Timelines & \textbf{0.317} & \textbf{0.952} & \textbf{0.916} & \textbf{0.475} & \textbf{0.661} & \textbf{0.649} & \textbf{0.892} & \textbf{0.654}& \textbf{0.739} & \textbf{0.524} & \textbf{0.890} & \textbf{0.613}\\
\hline

\end{tabular}
}
\label{tab:newone}
\end{table*}

\subsection{Model Comparison}
In the second set of experiments, we compare our best model (Full-Mode with shared tags, $P_n$=3, $h$=3) with the current state of the art topic tag recommendation techniques using the same metrics. We also establish a baseline using Latent Dirichlet allocation (LDA) method of tag prediction with an implementation similar to \cite{blei2003latent} ,\cite{hassan2018semantic}. 
We identified two state of the art models based on architectural and feature extraction differences. The first is HashRec \cite{wang-etal-2019-microblog} and the second is Attention-based Multimodal Neural Network Model for Hashtag Recommendation (AMNN) \cite{yuan2019attention}. HashRec is a hashtag recommendation model for social media text, specifically, conversational text on Twitter and Weibo. The model uses two encoders to encode the user posts and the conversation individually, followed by a bi-attention module to capture their interactions. The extracted features are further merged and fed into the hashtag decoder consisting of sequential Gated Recurrent Units (GRUs). We provide the first encoder with Profile Bios and the second encoder with a combination of prior posts, their keywords and shared tags from the neighborhood. We train the model to produce future tags sequentially. The default (Twitter) hyper-parameters and settings were used.  
AMNN is an attention based model for predicting tags from multi-modal data. The model uses two individual encoders, one for text and one for images consisting of bi-directional Long- Short Term Memory(LSTM) modules followed by an attention unit. The features are concatenated and fed to a sequential GRU model to predict tags. We modified the network to have two text encoders by removing the image encoder and the remainder of the network is unchanged. Similar to the previous model, the profile bios are provided to one of the encoders and the remaining input document is given to the next text encoder. We used the default hyper-parameters and settings to obtain results. The SOTA models are trained and tested on the same training set and test set used for our Bio-BERT (BBERT$_{Tg}$) model. \textbf{Table \ref{tab:comparison}} reports the results of this experiment. The best performance is in \textbf{bold} while the second best is \underline{underlined}.

\begin{table*}

\caption{Influence of Neighbourhood users in Input document with $P_n$=3 and full mode data accumulation, Averaged over 4 individual runs }
\label{tab:neigh}       
\center
\resizebox{\linewidth}{!}
{
\begin{tabular}{|c| c | c| c | c | c | c|c|c|c|c|c|c| }

\hline
 & \multicolumn{4}{|c|}{$K$=1} & \multicolumn{4}{|c|}{$K$=3} & \multicolumn{4}{|c|}{$K$=5} \\
\hline
Neighbourhood Users & Recall & Precision & Cosine Similarity & F1 &  Recall & Precision & Cosine Similarity & F1 &  Recall & Precision & Cosine Similarity & F1\\
\hline
$h$=0 (Original Full Model) & 0.296 & 0.874 & 0.909 & 0.442 & 0.590 & 0.584 & 0.891 & 0.586 & 0.681 & 0.402 & 0.887 & 0.505\\

$h$=1 & 0.302 & 0.898 & 0.911 & 0.451 & 0.601 & 0.597 & 0.891 & 0.598 & 0.712 & 0.437 & 0.889 & 0.541\\

$h$=2 & \textbf{0.319} & \textbf{0.957} & \textbf{0.916} &  \textbf{0.478} & 0.648 & 0.636 & 0.891 & 0.641 & 0.730 & 0.495 & 0.889 & 0.589\\
$h$=3 &0.317 & 0.952 & \textbf{0.916}& 0.475 &\textbf{0.661} & \textbf{0.649} & \textbf{0.892} & \textbf{0.654}  & \textbf{ 0.739} & \textbf{0.524} & \textbf{0.890} & \textbf{0.613}\\
$h$=4 &0.316 & 0.948 & \textbf{0.916} & 0.474 & 0.655 & 0.640 & 0.890 &0.647 & 0.733 & 0.501 & 0.877 & 0.595\\
$h$=5 &0.309 & 0.912 & 0.908 &0.461 & 0.652 & 0.638 & 0.888 &0.644 & 0.725 & 0.460 & 0.871 & 0.562\\
\hline

\end{tabular}
}
\end{table*}
\begin{table*}

\caption{Model Comparison with State of the art works and BBERT$_{Cl}$ }
\label{tab:comparison}       
\center
\resizebox{\linewidth}{!}
{
\begin{tabular}{|c| c | c| c | c | c | c|c|c|c|c|c|c| }

\hline
 & \multicolumn{4}{|c|}{$K$=1} & \multicolumn{4}{|c|}{$K$=3} & \multicolumn{4}{|c|}{$K$=5} \\
\hline

Model & Recall & Precision & Cosine Similarity & F1 &  Recall & Precision & Cosine Similarity & F1 &  Recall & Precision & Cosine Similarity & F1\\
\hline
LDA & 0.137 & 0.419 & 0.642 & 0.206 & 0.320 & 0.445 & 0.619 &0.372 & 0.480 & 0.203 & 0.606 & 0.285\\
AMNN & 0.264 & 0.794 & \underline{0.847} &0.396 & 0.585 & 0.529 & \underline{0.795} & 0.555 & 0.627 & 0.497 & 0.715 & 0.554\\

HashRec & \underline{0.278} & \underline{0.836} & 0.771 & \underline{0.417} & \underline{0.602} & \underline{0.563} & 0.736 & \underline{0.581} & \underline{0.681} & \underline{0.503} & \underline{0.773} & \underline{0.578}\\
BBERT$_{Cl}$ (Ours) & 0.266 & 0.797 & 0.834 & 0.398 & 0.536 & 0.499 & 0.718 &0.516 & 0.593 & 0.438 & 0.695 & 0.503\\
BBERT$_{Tg}$ (Ours) & \textbf{0.317} & \textbf{0.952} & \textbf{0.916} &\textbf{0.475} & \textbf{0.661} & \textbf{0.649} & \textbf{0.892} & \textbf{0.654} & \textbf{0.739} & \textbf{0.524} & \textbf{0.890} & \textbf{0.613}\\
\hline

\end{tabular}
}
\end{table*}

\begin{table}[h!]

\caption{Impact of Frequent Tags in Tag Prediction Model with $K$=3 (brackets indicate the change in absolute value)}

\label{tab:2tag}       
\center
\resizebox{\linewidth}{!}
{
\begin{tabular}{|c| c | c| c |c| }

\hline
  &\multicolumn{4}{|c|}{$K$=3} \\
\hline
 Model & Recall & Precision & Cosine Similarity & F1\\
\hline
LDA  & 0.320 & 0.445 & 0.619 & 0.372\\
BBERT$_{Tg}$  & 0.661 & 0.649 & 0.892 & 0.654\\
LDA $_{-2Tg}$ & 0.257 (-0.063) & 0.372 (-0.073) & 0.429 (-0.190) & 0.303 (-0.069)\\
BBERT$_{-2Tg}$  & 0.658 (-0.003) & 0.643 (-0.006) & 0.878 (-0.014) & 0.650 (-0.004)\\
\hline

\end{tabular}
}
\end{table}

\begin{table}

\caption{Qualitative Tag Comparison}

\label{tab:tag}       
\center
\resizebox{\linewidth}{!}
{
\begin{tabular}{|c|c| }
\hline
\multicolumn{2}{|c|}{  } \\
\multicolumn{2}{|c|}{Ground Truth Tags (Future Post):}\\
\multicolumn{2}{|c|}{Chemotherapy / Radiotherapy / Tarceva / Afatinib / Cancer and Tumors } \\

\hline
 {Model} &  {Predicted Tags}   \\
\hline
LDA  & Chemotherapy / Cancer / Lung / Treatment / Medication \\ 
HashRec & Chemotherapy / Cough / Cancer / Infection / Medication  \\
AMNN &  Chemotherapy / Tumor / Medication / Respiratory / Cough \\
BBERT$_{Cl}$ (Ours) &   Chemotherapy / Infection  / Cancer / Tumors / Cough \\
BBERT$_{Tg}$ (Ours) &   Chemotherapy / Radiotherapy  / Afatinib / Cancer and Tumors / Drug Trial \\
\hline

\end{tabular}
}
\end{table}

\subsection{Data Bias Experiments}\label{dbt}
Classification based models even with Bio-BERT embeddings do not perform as accurately as text generation models due to the imbalance in the prediction classes (tags). We show this by comparing a Bio-BERT classification (BBERT$_{Cl}$) model against our proposed tag generation model (BBERT$_{Tg}$). This is included in \textbf{Table \ref{tab:comparison}}. Very frequently occurring tags can skew results as they make it easier for models to guess the most frequent tag multiple times.  We checked for the robustness of LDA  and the BBERT$_{Tg}$ models by deleting the top two tags (\textit{Chemotherapy} and \textit{Cancer and Tumors}) from the train and test sets. This experiment is to measure the impact of tag frequency on the models. \textbf{Table \ref{tab:2tag}} shows the F1 scores when ($K$=3) tags are predicted with the two modified models. 
\section{Interpretation of Results} \label{resudisc}
Modeling information seeking behavior in a healthcare setting requires identifying key variables like disease similarity and history of interests. We designed our ablation studies to identify these factors and leverage the information to develop an accurate temporal tag prediction model. We observe that the amount of information presented to the BBERT$_{Tg}$ model largely affects the model's ability to understand topics of interest. From \textbf{Table \ref{tab:accu}}, we can see that using only the profile bios of the users yielded the lowest F1 score. This is due to low correlation between fixed content in the bio and the dynamic changes in search interest. Thus utilizing multiple data sources like the posts and their corresponding keywords help in improving predictions.  As anticipated, the number of previous posts in the input document causes variation in the tag prediction metrics. We notice that the F1 score is the highest when 2 previous posts along with the current post are used for future topic tag prediction. 
Improvements to the model come through collaborative filtering, an additional source of information regarding similar users in the database. In \textbf{Table \ref{tab:newone}}, experimenting with random neighbors yielded lower F1 scores at all levels indicating that including search tags from random users only caused the model to pay attention to noisy inputs. While on the other hand,
Bio-BERT features extracted from cleaned NER disease timelines yielded the cleanest set of neighbors who positively contributed to the tag prediction metrics. The search tags from neighbouring users helped the Bio-BERT model predict better as it is likely that they have queried about this topic in the past. 
An interesting observation from \textbf{Table \ref{tab:neigh}} is that the number of nearest users did not affect the cosine similarity significantly. We believe the reason for this is that since cosine similarity is calculated in the feature space, the user clusters are also defined in a similar feature space. Due to this, adding more users does not negatively affect cosine similarity. It did however point out that using more than 3 ($h>$3) neighbors could induce noise in the model due to uniqueness in the timelines. 

In \textbf{Table \ref{tab:comparison}}, we illustrate comparison of our proposed $BBERT_{Tg}$ model against the state of the art networks. Tag prediction models tend to perform well when the tags to be predicted depend only on the input document provided. However, understanding context and predicting the a future tag is a non-trivial task that leads to low F1 score with the HashRec and AMNN models. To overcome this, we pose the problem as a sentence generation task instead of the traditional classification task and obtain benchmark performance . We even use the Bio-BERT embeddings in a classification task ($BBERT_{Cl}$) to show that sentence generation performs better due to robustness to skewed tag distribution which is usually the case with social media based text data.  

Finally, we emphasize that our $BBERT_{Tg}$ model is robust to tag occurrence frequency and show that even when the 2 most frequent tags are removed, the performance does not change by much when compared to a traditional LDA model as shown in \textbf{Table \ref{tab:2tag}}.  

To show that our model is truly understanding the context, we show an example of generated tags for a future post by providing the previous posts, tags, profile-bio etc to all the models in comparison. The $future$ post \underline{\textbf{not}} provided to  the model was: 
\textit{ ``My partner was diagnosed with stage 4 lung cancer (nsclc) in June 2009. He has had Chemotherapy, Radiotherapy and Tarceva. He started taking the new unlicensed drug Afatinib yesterday. So far no side effects! He has been told that the side effects can be more severe than Tarceva. I really hope this drug works for him as we are fast running out of options."} Our BBERT$_{Tg}$ model was the only one to predict the tag \textit{Afatinib} and also understand that the patient is undergoing a drug trial. This is shown in \textbf{Table \ref{tab:tag}}.

\section{Summary and Discussion} \label{sumandisc}

In summary, our topic recommendation system, a hybrid system combining  
collaborative-filtering and content based ideas is designed for online health community users and utilises patient timeline similarity to group similar users. We use a pretrained Bio-BERT to perform the tasks of classification and sequential sentence generation. Our predictions of future topics yield accurate and contextual tags. We compare our model against models proposed for similar tasks like HashRec and AMNN and observe a superior performance. 


We have empirically presented the quantitative advantage of our methodology over existing models. This improvement in precision and recall metrics is attributed to two specific contributions- (1) An auxiliary sentence generation task for tag generation instead of tag classification and (2) Usage of ``disease timelines" to match similar users and perform collaborative filtering. Additionally, we also performed ablation studies to understand the importance of several factors like history and influence of similar users in the network. 
As a part of future work, we will focus on two specific aspects. First, we intend to improve the matching of similar users by leveraging interactions and graphs. Second, we plan to utilize the predicted tags to retrieve tailored medical articles from trusted sources like WebMD and Mayo clinic.

\bibliographystyle{IEEEtranS}
\bibliography{main}

\begin{thebibliography}{10}
\providecommand{\url}[1]{#1}
\csname url@samestyle\endcsname
\providecommand{\newblock}{\relax}
\providecommand{\bibinfo}[2]{#2}
\providecommand{\BIBentrySTDinterwordspacing}{\spaceskip=0pt\relax}
\providecommand{\BIBentryALTinterwordstretchfactor}{4}
\providecommand{\BIBentryALTinterwordspacing}{\spaceskip=\fontdimen2\font plus
\BIBentryALTinterwordstretchfactor\fontdimen3\font minus
  \fontdimen4\font\relax}
\providecommand{\BIBforeignlanguage}[2]{{%
\expandafter\ifx\csname l@#1\endcsname\relax
\typeout{** WARNING: IEEEtranS.bst: No hyphenation pattern has been}%
\typeout{** loaded for the language `#1'. Using the pattern for}%
\typeout{** the default language instead.}%
\else
\language=\csname l@#1\endcsname
\fi
#2}}
\providecommand{\BIBdecl}{\relax}
\BIBdecl

\bibitem{aceto2018role}
G.~Aceto, V.~Persico, and A.~Pescap{\'e}, ``The role of information and
  communication technologies in healthcare: taxonomies, perspectives, and
  challenges,'' \emph{Journal of Network and Computer Applications}, vol. 107,
  pp. 125--154, 2018.

\bibitem{barrett2016creating}
M.~Barrett, E.~Oborn, and W.~Orlikowski, ``Creating value in online
  communities: The sociomaterial configuring of strategy, platform, and
  stakeholder engagement,'' \emph{Information Systems Research}, vol.~27,
  no.~4, pp. 704--723, 2016.

\bibitem{beltagy2020longformer}
I.~Beltagy, M.~E. Peters, and A.~Cohan, ``Longformer: The long-document
  transformer,'' \emph{arXiv preprint arXiv:2004.05150}, 2020.

\bibitem{blei2003latent}
D.~M. Blei, A.~Y. Ng, and M.~I. Jordan, ``Latent dirichlet allocation,''
  \emph{the Journal of machine Learning research}, vol.~3, pp. 993--1022, 2003.

\bibitem{chan2019recurrent}
Y.-H. Chan and Y.-C. Fan, ``A recurrent bert-based model for question
  generation,'' in \emph{Proceedings of the 2nd Workshop on Machine Reading for
  Question Answering}, 2019, pp. 154--162.

\bibitem{day2019feasibility}
F.~C. Day, M.~Pourhomayoun, D.~Keeves, A.~F. Lees, M.~Sarrafzadeh, D.~Bell, and
  M.~A. Pfeffer, ``Feasibility study of an ehr-integrated mobile shared
  decision making application,'' \emph{International journal of medical
  informatics}, vol. 124, pp. 24--30, 2019.

\bibitem{deng2019collaborative}
X.~Deng and F.~Huangfu, ``Collaborative variational deep learning for
  healthcare recommendation,'' \emph{IEEE Access}, vol.~7, pp.
  55\,679--55\,688, 2019.

\bibitem{deng2017understanding}
Z.~Deng and S.~Liu, ``Understanding consumer health information-seeking
  behavior from the perspective of the risk perception attitude framework and
  social support in mobile social media websites,'' \emph{International journal
  of medical informatics}, vol. 105, pp. 98--109, 2017.

\bibitem{devlin2018bert}
J.~Devlin, M.-W. Chang, K.~Lee, and K.~Toutanova, ``Bert: Pre-training of deep
  bidirectional transformers for language understanding,'' \emph{arXiv preprint
  arXiv:1810.04805}, 2018.

\bibitem{ettinger2019nccn}
D.~S. Ettinger, D.~E. Wood, C.~Aggarwal, D.~L. Aisner, W.~Akerley, J.~R.
  Bauman, A.~Bharat, D.~S. Bruno, J.~Y. Chang, L.~R. Chirieac \emph{et~al.},
  ``Nccn guidelines insights: non--small cell lung cancer, version 1.2020:
  featured updates to the nccn guidelines,'' \emph{Journal of the National
  Comprehensive Cancer Network}, vol.~17, no.~12, pp. 1464--1472, 2019.

\bibitem{freitag2017beam}
M.~Freitag and Y.~Al-Onaizan, ``Beam search strategies for neural machine
  translation,'' \emph{arXiv preprint arXiv:1702.01806}, 2017.

\bibitem{9364676}
S.~Gao, M.~Alawad, M.~T. Young, J.~Gounley, N.~Schaefferkoetter, H.~J. Yoon,
  X.-C. Wu, E.~B. Durbin, J.~Doherty, A.~Stroup, L.~Coyle, and G.~Tourassi,
  ``Limitations of transformers on clinical text classification,'' \emph{IEEE
  Journal of Biomedical and Health Informatics}, vol.~25, no.~9, pp.
  3596--3607, 2021.

\bibitem{garg2020bae}
S.~Garg and G.~Ramakrishnan, ``Bae: Bert-based adversarial examples for text
  classification,'' in \emph{Proceedings of the 2020 Conference on Empirical
  Methods in Natural Language Processing}, 2020, pp. 6174--6181.

\bibitem{genes2018smartphone}
N.~Genes, S.~Violante, C.~Cetrangol, L.~Rogers, E.~E. Schadt, and Y.-F.~Y.
  Chan, ``From smartphone to ehr: a case report on integrating
  patient-generated health data,'' \emph{NPJ digital medicine}, vol.~1, no.~1,
  pp. 1--6, 2018.

\bibitem{guo2018cran}
L.~Guo, D.~Zhang, L.~Wang, H.~Wang, and B.~Cui, ``Cran: a hybrid cnn-rnn
  attention-based model for text classification,'' in \emph{International
  Conference on Conceptual Modeling}.\hskip 1em plus 0.5em minus 0.4em\relax
  Springer, 2018, pp. 571--585.

\bibitem{hassan2018semantic}
H.~A.~M. Hassan, G.~Sansonetti, F.~Gasparetti, and A.~Micarelli,
  ``Semantic-based tag recommendation in scientific bookmarking systems,'' in
  \emph{Proceedings of the 12th ACM Conference on Recommender Systems}, 2018,
  pp. 465--469.

\bibitem{9112671}
H.~Jelodar, Y.~Wang, R.~Orji, and S.~Huang, ``Deep sentiment classification and
  topic discovery on novel coronavirus or covid-19 online discussions: Nlp
  using lstm recurrent neural network approach,'' \emph{IEEE Journal of
  Biomedical and Health Informatics}, vol.~24, no.~10, pp. 2733--2742, 2020.

\bibitem{jiang2017user}
L.~Jiang and C.~C. Yang, ``User recommendation in healthcare social media by
  assessing user similarity in heterogeneous network,'' \emph{Artificial
  intelligence in medicine}, vol.~81, pp. 63--77, 2017.

\bibitem{joukes2019impact}
E.~Joukes, N.~F. de~Keizer, M.~C. de~Bruijne, A.~Abu-Hanna, and R.~Cornet,
  ``Impact of electronic versus paper-based recording before ehr implementation
  on health care professionals' perceptions of ehr use, data quality, and data
  reuse,'' \emph{Applied clinical informatics}, vol.~10, no.~02, pp. 199--209,
  2019.

\bibitem{kaageback2016word}
M.~K{\aa}geb{\"a}ck and H.~Salomonsson, ``Word sense disambiguation using a
  bidirectional lstm,'' \emph{arXiv preprint arXiv:1606.03568}, 2016.

\bibitem{kaur2018efficient}
H.~Kaur, N.~Kumar, and S.~Batra, ``An efficient multi-party scheme for privacy
  preserving collaborative filtering for healthcare recommender system,''
  \emph{Future Generation Computer Systems}, vol.~86, pp. 297--307, 2018.

\bibitem{kim2020patent}
J.~Kim, J.~Yoon, E.~Park, and S.~Choi, ``Patent document clustering with deep
  embeddings,'' \emph{Scientometrics}, pp. 1--15, 2020.

\bibitem{klavsnja2018enhancing}
A.~Kla{\v{s}}nja-Mili{\'c}evi{\'c}, M.~Ivanovi{\'c}, B.~Vesin, and Z.~Budimac,
  ``Enhancing e-learning systems with personalized recommendation based on
  collaborative tagging techniques,'' \emph{Applied Intelligence}, vol.~48,
  no.~6, pp. 1519--1535, 2018.

\bibitem{kruse2018health}
C.~S. Kruse and A.~Beane, ``Health information technology continues to show
  positive effect on medical outcomes: systematic review,'' \emph{Journal of
  medical Internet research}, vol.~20, no.~2, p. e8793, 2018.

\bibitem{kutner2006health}
M.~Kutner, E.~Greenburg, Y.~Jin, and C.~Paulsen, ``The health literacy of
  america's adults: Results from the 2003 national assessment of adult
  literacy. nces 2006-483.'' \emph{National Center for education statistics},
  2006.

\bibitem{lee2020biobert}
J.~Lee, W.~Yoon, S.~Kim, D.~Kim, S.~Kim, C.~H. So, and J.~Kang, ``Biobert: a
  pre-trained biomedical language representation model for biomedical text
  mining,'' \emph{Bioinformatics}, vol.~36, pp. 1234--1240, 2020.

\bibitem{li2021pretrained}
J.~Li, T.~Tang, W.~X. Zhao, and J.-R. Wen, ``Pretrained language models for
  text generation: A survey,'' \emph{arXiv preprint arXiv:2105.10311}, 2021.

\bibitem{liao2019news}
Y.-S. Liao, J.-Y. Lu, and D.-R. Liu, ``News recommendation based on
  collaborative semantic topic models and recommendation adjustment,'' in
  \emph{2019 International Conference on Machine Learning and Cybernetics
  (ICMLC)}.\hskip 1em plus 0.5em minus 0.4em\relax IEEE, 2019, pp. 1--6.

\bibitem{liu2019text}
Y.~Liu and M.~Lapata, ``Text summarization with pretrained encoders,'' in
  \emph{Proceedings of the 2019 Conference on Empirical Methods in Natural
  Language Processing and the 9th International Joint Conference on Natural
  Language Processing (EMNLP-IJCNLP)}, 2019, pp. 3730--3740.

\bibitem{luo2021deep}
X.~Luo, P.~Gandhi, S.~Storey, and K.~Huang, ``A deep language model for symptom
  extraction from clinical text and its application to extract covid-19
  symptoms from social media,'' \emph{IEEE Journal of Biomedical and Health
  Informatics}, vol.~26, no.~4, pp. 1737--1748, 2021.

\bibitem{9756911}
A.~Mahyari, P.~Pirolli, and J.~A. Leblanc, ``A deep recurrent neural network
  with user-profile attention-based real-time physical exercises recommendation
  system in mhealth,'' \emph{IEEE Journal of Biomedical and Health
  Informatics}, pp. 1--1, 2022.

\bibitem{mchugh2012interrater}
M.~L. McHugh, ``Interrater reliability: the kappa statistic,'' \emph{Biochemia
  medica}, vol.~22, no.~3, pp. 276--282, 2012.

\bibitem{melamud2016context2vec}
O.~Melamud, J.~Goldberger, and I.~Dagan, ``context2vec: Learning generic
  context embedding with bidirectional lstm,'' in \emph{Proceedings of the 20th
  SIGNLL conference on computational natural language learning}, 2016, pp.
  51--61.

\bibitem{miller2019leveraging}
D.~Miller, ``Leveraging bert for extractive text summarization on lectures,''
  \emph{arXiv preprint arXiv:1906.04165}, 2019.

\bibitem{morahan2000information}
J.~Morahan-Martin and C.~D. Anderson, ``Information and misinformation online:
  Recommendations for facilitating accurate mental health information retrieval
  and evaluation,'' \emph{CyberPsychology \& Behavior}, vol.~3, no.~5, pp.
  731--746, 2000.

\bibitem{nallapati2016sequence}
R.~Nallapati, B.~Xiang, and B.~Zhou, ``Sequence-to-sequence rnns for text
  summarization,'' 2016.

\bibitem{scikit-learn}
F.~Pedregosa, G.~Varoquaux, A.~Gramfort, V.~Michel, B.~Thirion, O.~Grisel,
  M.~Blondel, P.~Prettenhofer, R.~Weiss, V.~Dubourg, J.~Vanderplas, A.~Passos,
  D.~Cournapeau, M.~Brucher, M.~Perrot, and E.~Duchesnay, ``Scikit-learn:
  Machine learning in {P}ython,'' \emph{Journal of Machine Learning Research},
  vol.~12, pp. 2825--2830, 2011.

\bibitem{sahoo2019deepreco}
A.~K. Sahoo, C.~Pradhan, R.~K. Barik, and H.~Dubey, ``Deepreco: deep learning
  based health recommender system using collaborative filtering,''
  \emph{Computation}, vol.~7, no.~2, p.~25, 2019.

\bibitem{shu2018content}
J.~Shu, X.~Shen, H.~Liu, B.~Yi, and Z.~Zhang, ``A content-based recommendation
  algorithm for learning resources,'' \emph{Multimedia Systems}, vol.~24,
  no.~2, pp. 163--173, 2018.

\bibitem{singh2017tagme}
A.~Singh, N.~Nagwani, and S.~Pandey, ``Tagme: A topical folksonomy based
  collaborative filtering for tag recommendation in community sites,'' in
  \emph{Proceedings of the 4th Multidisciplinary International Social Networks
  Conference}, 2017, pp. 1--7.

\bibitem{song2019mass}
K.~Song, X.~Tan, T.~Qin, J.~Lu, and T.-Y. Liu, ``Mass: Masked sequence to
  sequence pre-training for language generation,'' \emph{arXiv preprint
  arXiv:1905.02450}, 2019.

\bibitem{tibshirani2001estimating}
R.~Tibshirani, G.~Walther, and T.~Hastie, ``Estimating the number of clusters
  in a data set via the gap statistic,'' \emph{Journal of the Royal Statistical
  Society: Series B (Statistical Methodology)}, vol.~63, no.~2, pp. 411--423,
  2001.

\bibitem{tran2018hashtag}
V.~C. Tran, D.~Hwang, and N.~T. Nguyen, ``Hashtag recommendation approach based
  on content and user characteristics,'' \emph{Cybernetics and Systems},
  vol.~49, no. 5-6, pp. 368--383, 2018.

\bibitem{tuarob2013automatic}
S.~Tuarob, L.~C. Pouchard, and C.~L. Giles, ``Automatic tag recommendation for
  metadata annotation using probabilistic topic modeling,'' in
  \emph{Proceedings of the 13th ACM/IEEE-CS joint conference on Digital
  libraries}, 2013, pp. 239--248.

\bibitem{vaswani2017attention}
A.~Vaswani, N.~Shazeer, N.~Parmar, J.~Uszkoreit, L.~Jones, A.~N. Gomez,
  {\L}.~Kaiser, and I.~Polosukhin, ``Attention is all you need,''
  \emph{Advances in neural information processing systems}, vol.~30, 2017.

\bibitem{wang2018dkn}
H.~Wang, F.~Zhang, X.~Xie, and M.~Guo, ``Dkn: Deep knowledge-aware network for
  news recommendation,'' in \emph{Proceedings of the 2018 world wide web
  conference}, 2018, pp. 1835--1844.

\bibitem{wang-etal-2019-microblog}
Y.~Wang, J.~Li, I.~King, M.~R. Lyu, and S.~Shi, ``Microblog hashtag generation
  via encoding conversation contexts,'' in \emph{Proceedings of the 2019
  Conference of the North {A}merican Chapter of the Association for
  Computational Linguistics: Human Language Technologies}.\hskip 1em plus 0.5em
  minus 0.4em\relax Association for Computational Linguistics, Jun. 2019, pp.
  1624--1633.

\bibitem{wang2020detecting}
Z.~Wang, Z.~Yin, and Y.~A. Argyris, ``Detecting medical misinformation on
  social media using multimodal deep learning,'' \emph{IEEE journal of
  biomedical and health informatics}, vol.~25, no.~6, pp. 2193--2203, 2020.

\bibitem{white2014health}
R.~W. White and E.~Horvitz, ``From health search to healthcare: explorations of
  intention and utilization via query logs and user surveys,'' \emph{Journal of
  the American Medical Informatics Association}, vol.~21, no.~1, pp. 49--55,
  2014.

\bibitem{yamada2020luke}
I.~Yamada, A.~Asai, H.~Shindo, H.~Takeda, and Y.~Matsumoto, ``Luke: deep
  contextualized entity representations with entity-aware self-attention,''
  \emph{arXiv preprint arXiv:2010.01057}, 2020.

\bibitem{yuan2019attention}
J.~Yuan, Y.~Jin, W.~Liu, and X.~Wang, ``Attention-based neural tag
  recommendation,'' in \emph{International Conference on Database Systems for
  Advanced Applications}.\hskip 1em plus 0.5em minus 0.4em\relax Springer,
  2019, pp. 350--365.

\bibitem{yue2021overview}
W.~Yue, Z.~Wang, J.~Zhang, and X.~Liu, ``An overview of recommendation
  techniques and their applications in healthcare,'' \emph{IEEE/CAA Journal of
  Automatica Sinica}, 2021.

\bibitem{zarzour2018new}
H.~Zarzour, Z.~Al-Sharif, M.~Al-Ayyoub, and Y.~Jararweh, ``A new collaborative
  filtering recommendation algorithm based on dimensionality reduction and
  clustering techniques,'' in \emph{2018 9th international conference on
  information and communication systems (ICICS)}.\hskip 1em plus 0.5em minus
  0.4em\relax IEEE, 2018, pp. 102--106.

\bibitem{zheng2018relationship}
M.~Zheng, H.~Jin, N.~Shi, C.~Duan, D.~Wang, X.~Yu, and X.~Li, ``The
  relationship between health literacy and quality of life: a systematic review
  and meta-analysis,'' \emph{Health and quality of life outcomes}, vol.~16,
  no.~1, pp. 1--10, 2018.

\bibitem{zhou2019improving}
J.~Zhou and C.~Wang, ``Improving cancer survivors’e-health literacy via
  online health communities (ohcs): A social support perspective,''
  \emph{Journal of Cancer Survivorship}, pp. 1--9, 2019.

\end{thebibliography}



\end{document}